\documentclass[10pt,twocolumn,letterpaper]{article}

\usepackage{cvpr}
\usepackage{times}
\usepackage{epsfig}
\usepackage{graphicx}
\usepackage{amsmath}
\usepackage{amssymb}
\usepackage{algorithmic}
\usepackage{graphics}
\usepackage{algorithm}
\usepackage{subfig}
\usepackage{color}
\usepackage{caption}
\usepackage[english]{babel}

\newcommand{\I}{\mathbf{I}}
\newcommand{\bO}{\mathbf{O}}
\newcommand{\bD}{\mathbf{D}}
\newcommand{\bC}{\mathbf{C}}
\newcommand{\bF}{\mathbf{F}}
\newcommand{\bx}{\mathbf{x}}
\newcommand{\bs}{\mathbf{s}}

\usepackage[pagebackref=true,breaklinks=true,colorlinks,bookmarks=false]{hyperref}

\cvprfinalcopy % *** Uncomment this line for the final submission

\def\cvprPaperID{XXX} % *** Enter the CVPR Paper ID here

\ifcvprfinal\fi
\begin{document}

%%%%%%%%% TITLE
\title{Learning by tracking: Siamese CNN for robust target association}
%\title{Learning to track pedestrians: Siamese CNN and Gradient Boosting-based target association for robust tracking}

\author{Laura Leal-Taix\'e\\
TU M\"unchen\\
Munich, Germany\\
%{\tt\small leal@geod.baug.ethz.ch}
% For a paper whose autho are all at the same institution,
\and
Cristian Canton Ferrer\\
Microsoft\\
Redmond (WA), USA\\
%{\tt\small ccrcanton@microsoft.com}
\and
Konrad Schindler\\
ETH Zurich\\
Zurich, Switzerland\\
%{\tt\small konrad.schindler@geod.baug.ethz.ch}
}

\maketitle
%\thispagestyle{empty}

%%%%%%%%% ABSTRACT
\begin{abstract}
\vspace{-0.3cm}
This paper introduces a novel approach to the task of data association within the context of pedestrian tracking, by introducing a two-stage learning scheme to match pairs of detections. First, a Siamese convolutional neural network (CNN) is trained to learn descriptors encoding local spatio-temporal structures between the two input image patches, aggregating pixel values and optical flow information. Second, a set of contextual features derived from the position and size of the compared input patches are combined with the CNN output by means of a gradient boosting classifier to generate the final matching probability. This learning approach is validated by using a linear programming based multi-person tracker showing that even a simple and efficient tracker may outperform much more complex models when fed with our learned matching probabilities. Results on publicly available sequences show that our method meets state-of-the-art standards in multiple people tracking.
\end{abstract}

\section{Introduction}

\par One of the big challenges of computer vision is scene
understanding from video. Humans are often the center of attention of
a scene, which leads to the fundamental problem of detecting and
tracking them in a video.
To track multiple people, \emph{tracking-by-detection} has emerged as
the preferred method. That approach simplifies the problem by dividing
it into two steps.
First, find probable pedestrian locations independently in each
frame. Second, link corresponding detections across time to form
trajectories.

\par The linking step, called \emph{data association} is a difficult
task on its own, due to missing and spurious detections, occlusions, and
targets interactions in crowded environments.
To address these issues, research in this area has produced more and
more complex models: global optimization methods based on network flow
\cite{berclaztpami2011,zhangcvpr2008}, minimum cliques
\cite{zamireccv2012} or discrete-continuous CRF inference
\cite{andriyenkocvpr2011}; models of pedestrian interaction with
social motion models \cite{lealiccv2011,pellegriniiccv2009};
integration of additional motion cues such as dense point trajectories
\cite{choiiccv2015,fragkiadakieccv2012}; and person re-identification
techniques to improve appearance models
\cite{kimiccv2015,kuocvpr2011}.
Even though the models became progressively more sophisticated, the
underlying descriptors, which are used to decide whether two
detections belong to the same trajectory, remained quite simple and
struggle in challenging scenarios (\eg, crowds, frequent occlusions,
strong illumination effects).
%
%\textcolor{red}{Careful: out of (I think) Nevatias lab, there was some
%  work with online discriminative learning to attack the problem.}
%: appearance \can{AddRef}, optical flow \can{AddRef} or simple motion models \can{AddRef}. 

\begin{figure}[!t]
\begin{center}
\includegraphics[width=\linewidth]{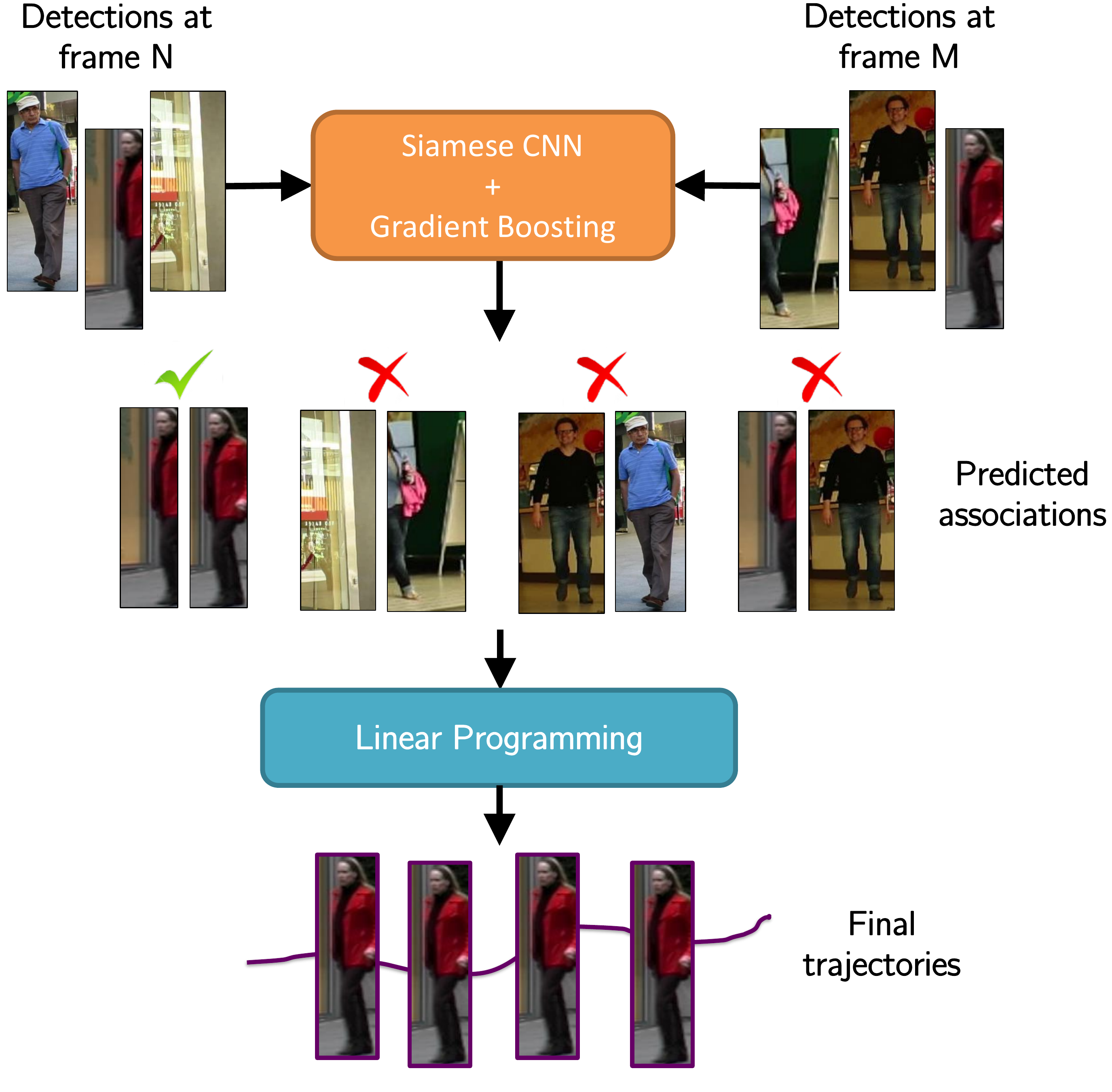}
\end{center}
   \caption{Multiple object tracking with learned detection associations.}
\label{fig:teaser}
\end{figure}
 
\par Recently, larger amounts of annotated data have become available
and, with the help of these data, convolutional neural networks (CNNs)
that learn feature representations as part of their training have
outperformed heuristic, hand-engineered features in several vision
problems \cite{krizhevskyImageNet}. Here, we adapt the CNN philosophy to multi-person tracking. In order to circumvent manual feature design for data association, we propose
to learn the decision whether two detections belong to the same trajectory. 
Our learning framework has two stages: first, a CNN in Siamese twin
architecture is trained to assess the similarity of two equally sized
image regions; second, contextual features that capture the relative
geometry and position of the two patches of interest are combined with
the CNN output to produce a final prediction, in our case using
gradient boosting (GB).
%In our case, we stack pixel and optical flow values as the input to
%our system thus allowing the CNN to learn joint appearance and motion
%templates.
Given the learned, pairwise data association score we construct a
graph that links all available detections across frames, and solve the
standard Linear Programming (LP) formulation of multi-target tracking.
We show that this simple and efficient linear tracker -- in some sense
the ``canonical baseline'' of modern multi-target tracking --
outperforms much more complex models when fed with our learned edge
costs.

\begin{figure*}[!ht]
\begin{center}
\includegraphics[width=\linewidth]{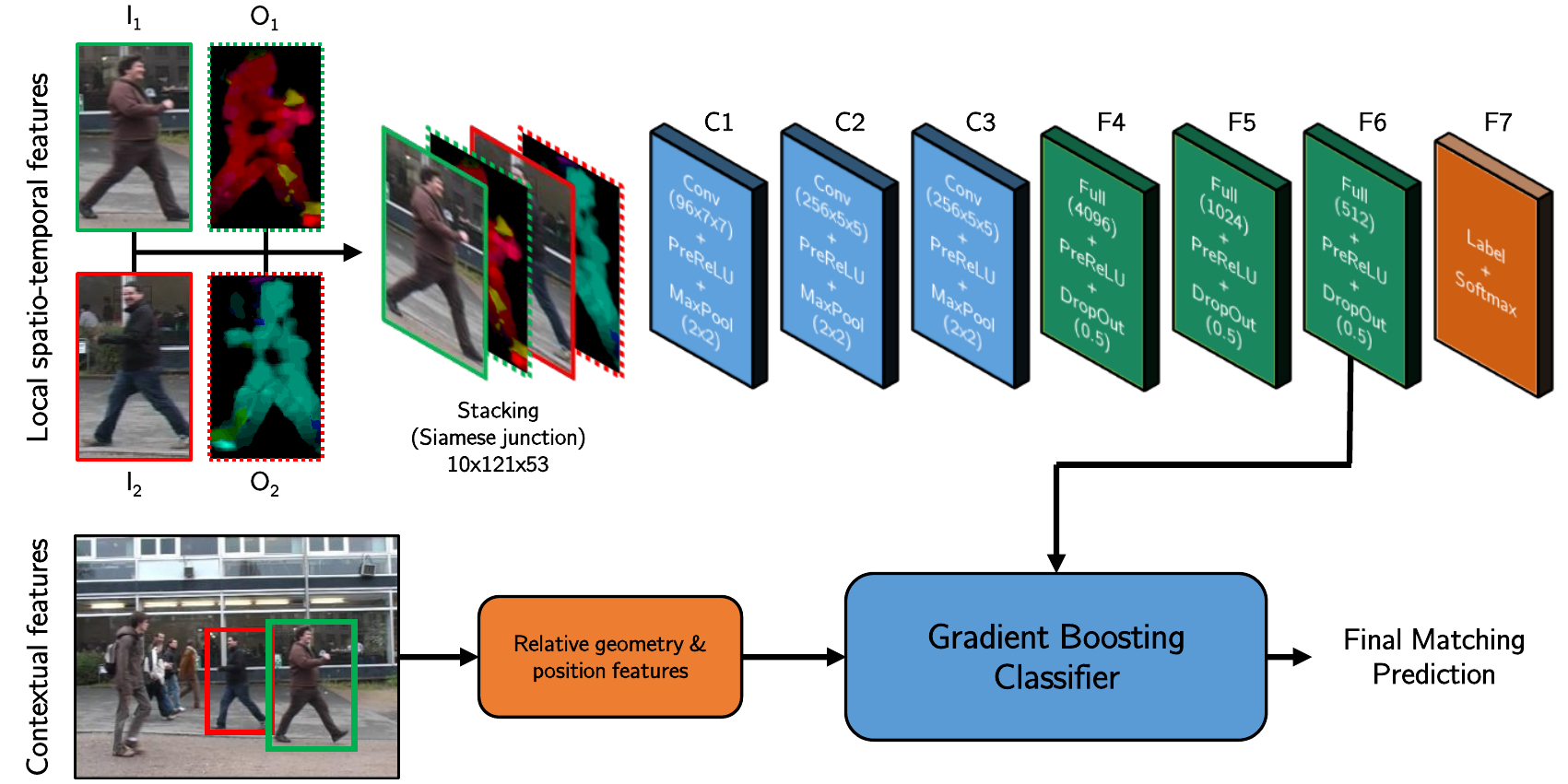}
\end{center}
\vspace{-0.25cm}
   \caption{Proposed two-stage learning architecture for pedestrian detection matching.}
\label{fig:learningArchitecture}
\end{figure*}

\subsection{Contributions}

This paper presents three major contributions to the pedestrian tracking task:

\begin{itemize}
\item{Within the context of tracking, we introduce a novel learning perspective to the data association problem.}
\item{We propose to use a CNN in a Siamese configuration to estimate the likelihood that two pedestrian detections belong to the same tracked entity. In the presented CNN architecture, pixel values and optical flow are combined as a multi-modal input.}\vspace{-0.1cm}
\item{We show that formulating data association with a linear optimization model outperform complex models when fed with accurate edge costs.}\vspace{-0.1cm}
\end{itemize}

\subsection{Related work}

%\lau{I need to rewrite Related work, do not work on it :)}
% A bit on detection too

\paragraph{Multi-person tracking.}

\par Multi-person tracking is the input for a number of computer vision
applications, such as surveillance, activity recognition or autonomous
driving. Despite the vast literature on the topic
\cite{Luo:2014:arXiv}, it still remains a challenging problem,
especially in crowded environments where occlusions and false
detections are common.
Most modern methods use the tracking-by-detection paradigm, which
divides the task into two steps: detecting pedestrians in the scene
\cite{dollarpami2014,partbased,galltpami2011}, and linking those
detections over time to create trajectories.
A common formalism is to represent the problem as a graph, where each
detection is a node, and edges indicate a possible link. The data
association can then be formulated as maximum flow
\cite{berclaztpami2011} or, equivalently, minimum cost problem
\cite{jiangcvpr2007,lealiccv2011,pirsiavashcvpr2011,zhangcvpr2008},
both efficiently solved to (near-)global optimality with LP, with a
superior performance compared to frame-by-frame \cite{khantpami2005}
or track-by-track \cite{berclazcvpr2006} methods.
Alternative formulations typically lead to more involved optimization
problems, including minimum cliques \cite{zamireccv2012} or
general-purpose solvers like MCMC \cite{yucvpr2007}. There are also
models that represent trajectories in continuous space and use
gradient-based optimization, sometimes alternating with discrete
inference for data association \cite{andriyenkocvpr2011}.

A recent trend is to design ever more complex models, which include
further vision routines in the hope that they benefit the tracker,
including reconstruction for multi-camera sequences
\cite{lealcvpr2012,wucvpr2011}, activity recognition
\cite{choieccv2012} and segmentation \cite{milancvpr2015}.
In general, the added complexity seems to exhibit diminishing returns,
at significantly higher computational cost.

\par Other works have focused on designing more robust features to
discriminate pedestrians. Color-based appearance models are common
\cite{kimiccv2015}, but not always reliable, since people can wear
very similar clothes, and color statistics are often contaminated by
the background pixels and illumination changes.
Kuo \emph{et al.} \cite{kuocvpr2011}, borrow ideas from person re-identification and
adapt them to ``re-identify'' targets during tracking.
In \cite{yangcvpr2012}, a CRF model is learned to better distinguish 
pedestrians with similar appearance. 
A different line of attack is to develop sophisticated motion models
in order to better predict a tracked person's location, most notably
models that include interactions between nearby people
\cite{andriyenkocvpr2011,choieccv2010,lealiccv2011,pellegriniiccv2009,scovannericcv2009,
  yamaguchicvpr2011}.
A problem of such models is that they hand-craft a term for each
external influence (like collision avoidance, or walking in
groups). This limits their applicability, because it is difficult to
anticipate all possible interaction scenarios.
The problem can be to some degree alleviated by learning the motion
model from data \cite{lealcvpr2014}, although this, too, only works if
all relevant motion and interaction patterns are present in the
training data.
Moreover, the motion model does not seem to be an important bottleneck
in present tracking frameworks. By and large, more powerful dynamic
models seem to help only in a comparatively small number of
situations, while again adding complexity.

\paragraph{Measuring similarity with CNNs.}

\par Convolutional architectures have become the method of choice for
end-to-end learning of image representations. In relation to our problem, they have also been remarkably successful
in assessing the similarity of image patches for different tasks such
as optical flow
estimation \cite{flownet2015}, face verification \cite{deepFace2014},
and depth estimation from multiple viewpoints \cite{deepStereo2015,zagoruyko2015,Zbontar2015}.

In the context of tracking, CNNs have been used to model appearance
and scale variations of the target \cite{fan2010human}. Recently,
several authors employ them to track via online learning, by
continuously fine-tuning a pre-trained CNN model
\cite{Chen2015,Li2015,Wang2015}.

\section{Learning to associate detections}

\par Our tracking framework is based on the paradigm of tracking-by-detection, i.e. firstly, we run a detector through the sequences, and secondly, we link the detections to form trajectories. We propose to address the data association problem by learning a model to predict whether two detections belong to the same trajectory or not.
We use two sets of features derived from the pedestrian detections to be compared. First, \emph{local spatio-temporal features} learnt using a CNN and, second, \emph{contextual features} encoding the relative geometry and position variations of the two detections. Finally, both sets of features are combined using a GB classifier \cite{Friedman2002GBM} to produce the final prediction (see Fig.\ref{fig:learningArchitecture}). Decoupling local and global features processing and ensembling them in a later stage allows understanding the contribution of each factor plus adding robustness to the prediction \cite{DielemanPlankton2015,Wang2014}. 

\subsection{CNN for patch similarity}
\label{firststep}

\par
\begin{figure}[!t]
\subfloat[Cost function]{%
  \includegraphics[width=0.3\linewidth]{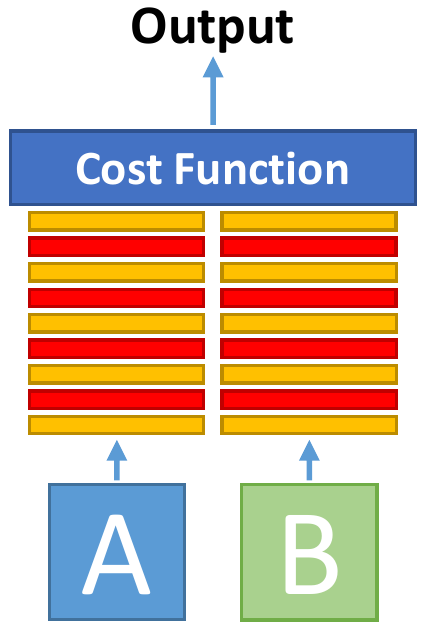}
}
\hfill
\subfloat[In-network]{%
  \includegraphics[width=0.3\linewidth]{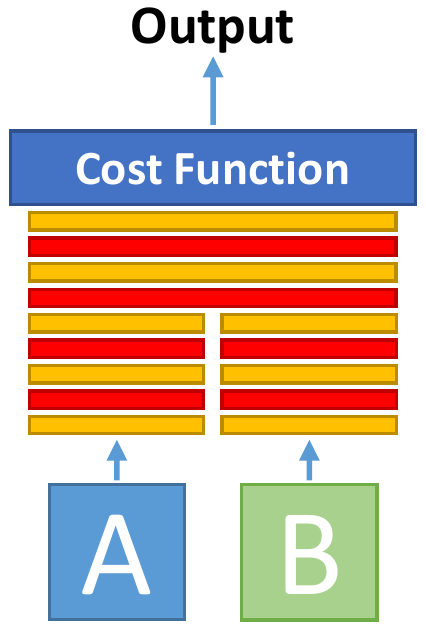}
}
\hfill
\subfloat[Input stacking]{%
  \includegraphics[width=0.3\linewidth]{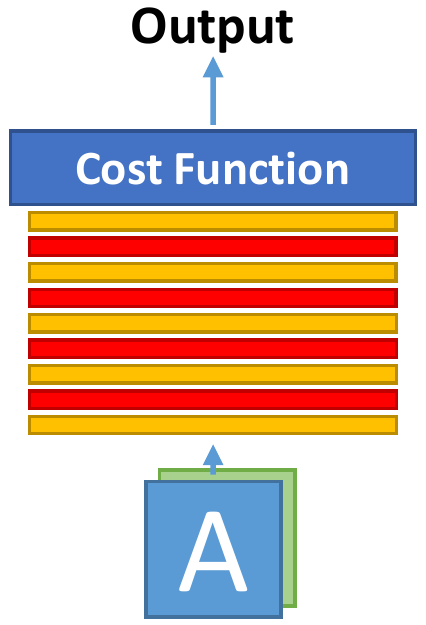}
}
\caption{Siamese CNN topologies}
\label{fig:SiameseCNNTopologies}
\end{figure}

 A common denominator when comparing two image patches using CNNs are Siamese architectures where two inputs are processed simultaneously by several layers with shared weights (convolutional and/or fully connected) that eventually merge at some point in the network. Siamese CNN topologies can be grouped under three main categories, depending on the point where the information from each input patch is combined (see Fig.\ref{fig:SiameseCNNTopologies}):

\begin{itemize}
\item \textbf{Cost function.} Input patches are processed by two parallel branches featuring the same network structure and weights. Finally, the top layers of each branch are fed to a cost function \cite{chopra2006,deepFace2014} that aims at learning a manifold where different classes are easily separable.
\item \textbf{In-network.} In this case, the top layers of the parallel branches processing the two different inputs are concatenated and some more layers are added on top of that \cite{flownet2015,Zbontar2015}. Finally, the standard softmax log-loss function is employed.
\item \textbf{Joint data input.} The two input patches are stacked together forming a unified input to the CNN \cite{flownet2015}. Again, the softmax log-loss function is used here.
\end{itemize}

\par While the two first approaches have yield good results in classification applications, the best performance for tasks involving comparison of detailed structures is obtained with the joint data input strategy. As pointed out by \cite{zagoruyko2015} and further corroborated by \cite{flownet2015}, jointly using information from both patches from the first layer tends to deliver a better performance. In order to verify this hypothesis within the scope of the tracking problem, we trained a Siamese network using the contrastive loss function \cite{Chopra2005}:
$$E = \frac{1}{2N} \sum\limits_{n=1}^N \left(y\right) d + \left(1-y\right) \max \left(\tau-d, 0\right),$$
where $ d = \left| \left| a_n - b_n \right| \right|_2^2$, being $a_n$ and $b_n$ the $L2$ {normalized} responses of the top fully connected layer of the parallel branches processing each input image, and $\tau=0.2$ is the separation margin and $y$ the label value encoded as $0$ or $1$. The topology of the CNN network has been the same all through the paper and shown in Fig.\ref{fig:learningArchitecture}. Our early experiments, showed a relative 8\% AUC increase of the joint data input case over the best performing model from the other two topologies, given a fixed number of parameters.
\newpage
\textbf{Architecture.} The proposed CNN architecture takes as input four sources of information: the pixel values in the normalized LUV color format for each patch to be compared, $\I_1$ and $\I_2$, and the corresponding $x$ and $y$ components of their associated optical flow \cite{opencvOF}, $\bO_1$ and $\bO_2$. These four images are resized to a fixed size of $121$x$53$ and stacked depth-wise to form a multi-modal 10-channel data blob $\bD$ to be fed to the CNN. In order to improve robustness against varying light conditions, for each luma channel L of both $\I_1$ and $\I_2$ we perform a histogram equalization and a plane fitting, as introduced in \cite{Cha2014}.

\par The input data is processed first by three convolutional layers, $\bC_{1,2,3}$, each of them followed by a PreReLU non-linearity \cite{kaiming2015} and a max-pooling layer that renders the net more robust to miss alignments within the components of $\bD$. Afterwards, four fully connected layers, $\bF_{4,5,6,7}$, aim at capturing correlations between features in distant parts of the image as well as cross-modal dependencies, i.e. pixel-to-motion interactions between $\I_{1,2}$ and $\bO_{1,2}$. The output of the last fully-connected layer is fed to a binary softmax which produces a distribution over the class labels (match/no match). The output of layer $\bF_6$ in the network will be used as our raw patch matching representation feature vector to be fed to the second learning stage.

\textbf{Training data generation.} Pedestrian detections proposed using \cite{dollarpami2014} are generated for each frame and associations between detections are provided across frames during the training phase. On one hand, positive examples, i.e. pairs of detections corresponding to target $m$, $(\I^m_{t},\I^m_{t-k})$, $1\leq k<N$, are directly generated from the ground truth data, with a maximum rewind time of $N=15$. On the other hand, negative examples are generated by either pairing two true detections with belonging to different people, a true detection with a false positive or two false positive detections; in order to increase the variety of data presented to the CNN, we enlarged the set of false positives by randomly selecting patches from the image of a given aspect ratio that do not overlap with true positive detections. By generating these random false positives, the CNN does not overfit to the specific type of false positives generated by the employed pedestrian detector thus increasing its capacity of generalization. 

\textbf{Learning.} We trained the proposed CNN as a binary classification task, employing the standard back-propagation on feed-forward nets by stochastic gradient descent with momentum. The mini-batch size was set to 128, with an equal learning rate for all layers set to $0.01$, sequentially decreased every $1.5$ epochs by a factor $10$, finally reaching $10^{-4}$. Layer weight were initialized following \cite{kaiming2015} and we trained our CNN on a Titan GPU X for 50 epochs. The Lasagne/Theano framework was employed to run our experiments.

\textbf{Data augmentation.} Even if the available training data is fairly large, pairs of pedestrian detections tend not to have a large range of appearances stemming from the fact that the number of distinct people in the training corpus is limited. Adding variety to the input data during the training phase is a widely employed strategy to reduce overfitting and improve generalization of CNNs \cite{DielemanPlankton2015,Dieleman2015,krizhevskyImageNet}. In our particular case, we have randomly added geometric distortions (rotation, translation, skewing, scaling and vertical flipping) as well as image distortions (Gaussian blur, noise and gamma). These transformations are applied independently for each of the two input patches but only allowing small relative geometric transformations between them (with the exception of vertical flipping that is applied to both images, when chosen). Since all these transformation are performed directly on GPU memory, the augmentation complexity cost is negligible.

\subsection{Evidence aggregation with gradient boosting}

\par The softmax output of the presented Siamese CNN might be used directly for pedestrian detection association but the accuracy would be low since we are not taking into account \emph{where} and \emph{when} these detections originated in the image. Therefore, the need for a set of contextual features and a higher order classifier to aggregate all this information.

\par Given two pedestrian detections at different time instants, $\I_{t_1}$ and $\I_{t_2}$, encoded by its position $\bx=(x,y)$ and dimensions $\bs=(w,h)$, we define our contextual features as: the relative size change, ${(\bs_1-\bs_2)/(\bs_1+\bs_2)}$, the position change, ${(\bx_1-\bx_2)}$, and the relative velocity between them, ${(\bx_1-\bx_2)/(t_2-t_1)}$. 

\par Combining the local and contextual sets of features is carried out using gradient boosting (GB) \cite{Friedman2002GBM}. To avoid overfitting on the GB, CNN predictions for each of the train sequences are generated in a leave-one-out fashion following the stacked generalization concept introduced in \cite{Wolpert92}. Finally, the GB classifier is trained by concatenating the CNN and contextual features. In our case, we trained the GB classifier using 400 trees using the distributed implementation presented in \cite{chen2015xgboost}.
\newpage
% \begin{figure}
% \begin{center}
% \includegraphics[width=\linewidth]{figures/ROC.pdf}
% \end{center}
% \caption{Performance accuracy for the Siamese CNN and the full two-stage learning approach.}
% \label{fig:ROC}
% \end{figure}
\section{Tracking with Linear Programming}
\label{TrackingLP}

In this section, we present the tracking framework where we incorporate the score defined in the previous section in order to solve the data association problem. 

Let $\mathcal{D}=\{{\bf d}^t_i\}$ be a set of object detections with ${{\bf d}^t_i=(x,y,t)}$, where $(x,y)$ is the 2D image position and $t$ defines the time stamp.
A trajectory is defined as a list of ordered object detections $T_k=\{{{\bf d}^{t_1}_{k_1}},{{\bf d}^{t_2}_{k_2}}, \cdots , {{\bf d}^{t_N}_{k_N}}\}$, and the goal of multiple object tracking is to find the set of trajectories $\mathcal{T}*=\{T_k\}$ that best explains the detections $\mathcal{D}$.
This can be expressed as a Maximum A-Posteriori (MAP) problem and directly mapped to a Linear Programming formulation, as detailed in \cite{lealiccv2011,zhangcvpr2008}. 

The data association problem is therefore defined by a linear program with objective function:
\begin{align}
\mathcal{T}*&=\underset{\mathcal{T}}{\operatorname{{\bf argmin}}}  \sum_{i} C_\textrm{in}(i)f_\textrm{in}(i)  + \sum_{i} C_\textrm{out}(i)f_\textrm{out}(i) \nonumber \\
&+ \sum_{i} C_\textrm{det}(i)f(i)  +\sum_{i,j}  C_\textrm{t}(i,j) f(i,j) 
\label{LP}
\end{align}
subject to edge capacity constraints, flow conservation at the nodes and exclusion constraints.

The costs $C_\textrm{in}$ and $C_\textrm{out}$ define how probable it is for a trajectory to start or end. The detection cost $C_\textrm{det} (i)$ is linked to the score that detection $i$ was given by the detector. Intuitively, if the score $s_i$ is very high, the cost of the edge should be very negative, so that flow will likely pass through this edge, including the detection $i$ in a trajectory.
We normalize the costs $s_i=[0,1]$ for a sequence, and define the detection cost as:
\begin{align}
C_\textrm{det}(i) =
\begin{cases}
 \frac{-s_i}{V_{\textrm{det}}} +1  &\mbox{ if } s_i < V_{\textrm{det}} \\
 \frac{-s_i+1}{1-V_{\textrm{link}}}-1  &\mbox{ if } s_i \geq V_{\textrm{det}}
\end{cases}
\label{detcost}
\end{align}
If we set, for example, $V_{\textrm{det}}=0.5$, the top half confident detections will correspond to edges with negative cost, and will most likely be used in some trajectory. By varying this threshold, we can adapt to different types of detectors that have different rates of false positives.

The cost of a link edge depends only on the probability that the two detections $i$ and $j$ belong to the same trajectory, as estimated by our classifier:
\begin{align}
C_\textrm{t}(i,j) =
\begin{cases}
 \frac{-s^{\textrm{RF}}_{i,j}}{V_{\textrm{link}}} +1  &\mbox{ if } s^{\textrm{RF}}_{i,j} < V_{\textrm{link}} \vspace{0.2cm} \\ 
 \frac{-s^{\textrm{RF}}_{i,j}+1}{1-V_{\textrm{link}}}-1  &\mbox{ if } s^{\textrm{RF}}_{i,j} \geq V_{\textrm{link}}
\end{cases}
\label{linkcost}
\end{align}
\normalsize

Note in Eq. \eqref{LP}, that if all costs are positive, the trivial solution will be zero flow. A trajectory is only created if its total cost is negative. 
We define detection costs to be negative if we are confident that the detection is a pedestrian, while transition costs are negative if our classifier is very confident that two detections belong to the same trajectory. 
We control with $V_{\textrm{det}}$ and $V_{\textrm{link}}$ the percentage of negative edges that we want in the graph. 
The in/out costs, on the other hand, are positive and they are used so that the tracker does not indiscriminately create many trajectories. 
Therefore, a trajectory will only be created if there is a set of confident detections and confident links whose negative costs outweigh the in/out costs. 
$C_\textrm{in}=C_\textrm{out}$, $V_{\textrm{det}}$ and $V_{\textrm{link}}$ are learned from training data as discussed in the next section.

The Linear Program in Eq. \eqref{LP} can be efficiently solved using Simplex \cite{lealiccv2011} or k-shortest paths \cite{berclaztpami2011}. 
Note, that we could use any other optimization framework, such as maximum cliques \cite{zamireccv2012}, or Kalman filter \cite{pellegriniiccv2009} for real-time applications.

\section{Experimental results}

\par This section presents the results validating the efficiency of the proposed learning approach to match pairs of pedestrian detections as well as its performance when creating trajectories by means of the aforementioned linear programming tracker. In order to provide comparable results with the rest of the state-of-the-art methods, we employed the large MOTChallenge \cite{motchallenge:arxiv:2015} dataset, a common reference when addressing multi-object tracking problems. 
%All results are done on the MOTChallenge \cite{motchallenge:arxiv:2015} multi-target tracking benchmark. 
It consists of 11 sequences for training, almost 40,000 bounding boxes, and 11 sequences for testing, over 60,000 boxes, comprising sequences with moving and static cameras, dense scenes, different viewpoints, etc. 
%\lau{Here we add the MOT16 data if we have it!}

% We present a thorough analysis of the behavior of the learned model, several variants of the proposed method as well as a baseline of manually generated features. Here is a description of the models compared: \\

\subsection{Detection matching}

\begin{figure}
\begin{center}
\includegraphics[width=\linewidth]{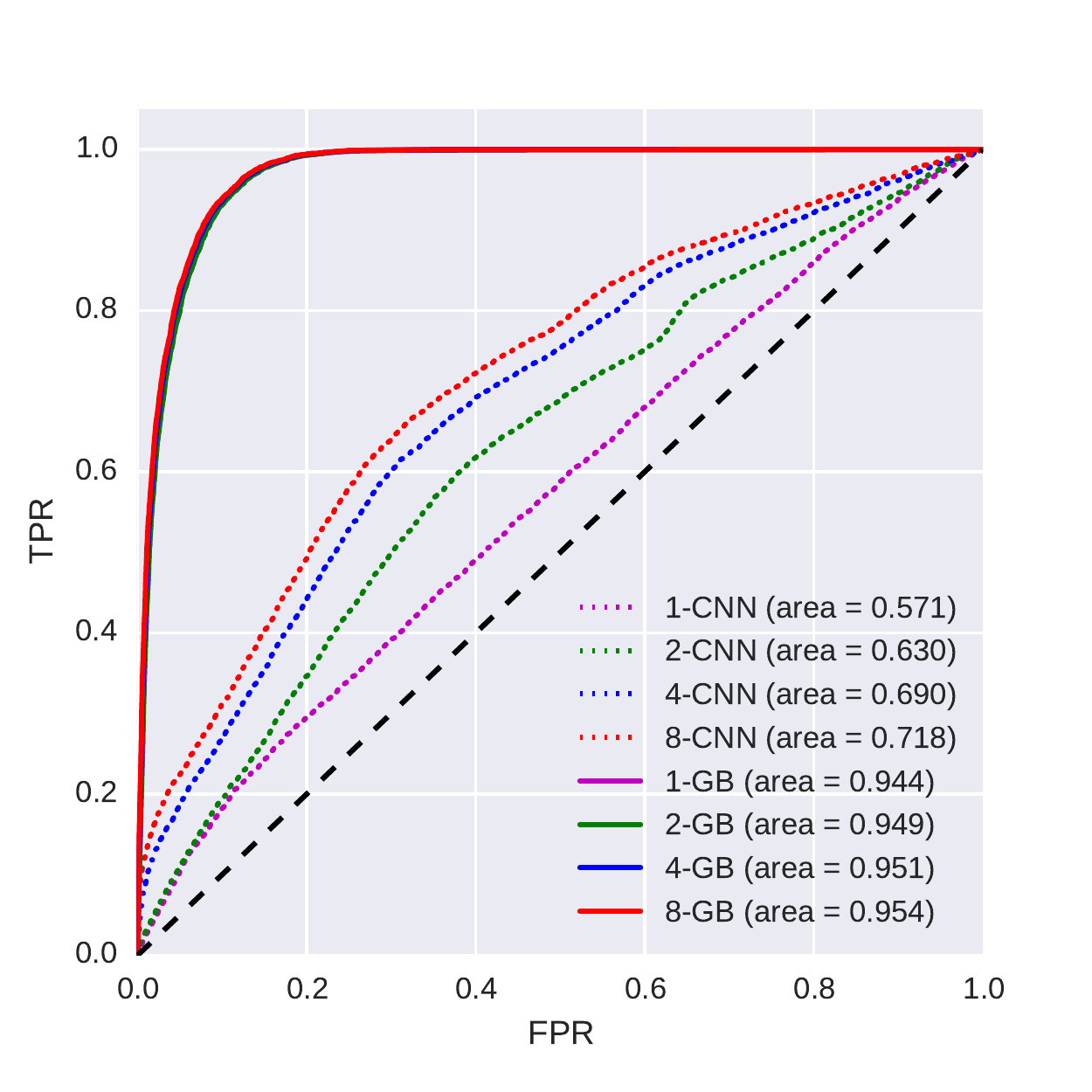}
\end{center}
\caption{Performance accuracy for the Siamese CNN and the full two-stage learning approach (CNN+GB), when using an oversampling of 8,4,2 and 1 per pair at the input.}
\label{fig:ROC}
\end{figure}

\par We first evaluate the performance of the proposed learning approach when predicting the probability of two detections belonging to the same trajectory by means of the ROC curve computed on the training data of MOT15 \cite{motchallenge:arxiv:2015}, as shown in Fig.\ref{fig:ROC}. Two result groups are depicted: first, when only using the CNN classifier (best AUC: $0.718$) and, second, when using the two stage CNN+GB classifier (best AUC: $0.954$); the later yielding to a relative $41$\% increase in classification performance. Oversampling the image (1,2,4 and 8 fixed locations) and averaging their predictions proved to deliver a significant improvement, specially for the CNN part of the end-to-end system. However, the impact of oversampling in the CNN+GB classifier is less relevant hence it may be avoided to reduce the overall computation load.

%\lau{The next paragraph is not clear imo. What is this, the way you compute the ROC? (In this case it should go before the previous paragraph). What does it mean that you oversample the image regions?}

%\par We compute the prediction of two patches belonging to the same pedestrian by randomly oversampling the corresponding image regions and we verify that, by increasing the number of samples, accuracy improves.

%For our experiments we selected an oversampling of 8. However, when evaluating the end-to-end system, we found out that contextual features are critical to reduce FPR. 
%As we can see, the two-stage learning with gradient boosting (GB) is able to effectively combine the two sources of information, yielding  a final AUC of 0.95.

%\par Performance of the two-stage learning approach for data association has been addressed by means of the ROC curve computed on the training data of MOT15 \cite{motchallenge:arxiv:2015}, as shown in Fig.\ref{fig:ROC}. Regarding the CNN, we compute the prediction of two patches belonging to the same pedestrian by randomly oversampling the corresponding image regions and we verified that, by increasing the number of samples, accuracy improves. For our experiments we selected an oversampling of 8. However, when evaluating the end-to-end system, we found out that contextual features are critical to reduce FPR. Finally, GB, when trained with the right parameters, is able to effectively combine the two sources yielding to an AUC of 0.95.

\par An analysis of the ROC curve on the MOT15 training data allowed us to find the operation point, i.e. probability threshold $V_{\textrm{link}}$ within the linear programming tracking, that would maximize its accuracy. In our case, we set ${V_{\textrm{link}} = 0.35}$, after cross-validation.

%\subsection{Analysis of different architectures}
%\label{arch-exps}

%\lau{Here we should have results for 4 variants}:
%\begin{enumerate}
%\item Two parallel networks, only image info
%\item Two parallel networks, with OF info
%\item Single network where we concatenate the input, only image info
%\item Single network where we concatenate the input, with OF info
%\end{enumerate}

%This should all be done with the training data of MOT15. The best performing network then goes to the next section on the test set.

\subsection{Multiple people tracking}
\label{mtt-exps}

\noindent{\bf Evaluation metrics.} To evaluate multiple-object tracking performance, we used CLEAR MOT metrics \cite{Bernardin:2008:CLE}, tracking accuracy (TA) and precision (TP). TA incorporates the three major error types (missing recall, false alarms and identity switches (IDsw)) while TP is a measure for the localization
error, where 100\% again reflects a perfect alignment of the output tracks and the ground truth. 
There are also two measures taken from \cite{Li:2009:CVPR} which reflect the temporal coverage of true trajectories by the tracker: mostly tracked (MT, $> 80\%$ overlap) and mostly lost (ML, $< 20\%$). 
We use only publicly available detections and evaluation scripts provided in the benchmark \cite{motchallenge:arxiv:2015}.\\

\noindent{\bf Determining optimal parameters.} As discussed before, the LP parameter $V_{\textrm{link}}=0.35$ is given by the operation point of the ROC curve. The other LP parameters, $C_\textrm{in}=C_\textrm{out}$, $V_{\textrm{det}}$ are determined by parameter sweep with cross-validation on the training MOT15 data in order to obtain the maximum tracking accuracy.\\

\noindent{\bf Baselines.} We compare to two tracking methods based on Linear Programming. The first is using only 2D distance information as feature (LP2D), the second \cite{lealcvpr2014} is learning to predict the motion of a pedestrian using image features (MotiCon). This comparison is specially interesting, since the optimization structure for all methods is based on Linear Programming, and the only factor that changes is the way the edge costs are computed. In this way, we can see the real contribution of our proposed learn-based costs. As it can be seen in Table \ref{tab:motcha-seqs}, the results indicate that our learned data association costs are more accurate, and that this better low-level evidence is the key factor driving the performance improvement.

\begin{table}[!t]
\begin{center}
\small
 \begin{tabular}{l | l|p{0.55cm}p{0.55cm}p{0.55cm}p{0.55cm}p{0.55cm}}
   Dataset & Method   & TA & TP & MT & ML  &  IDsw \\ \hline

&     MotiCon & 58.2 & 70.8 & 23.1 & 15.4 & 403  \\
  TUD-   & LP2D  & 49.5 & 74.1 & 15.4 & 15.4 & 48  \\
 Crossing & Proposed & 73.7 & 73.0 & 69.2 & 15.4 & 197 \\\hline
  
&    MotiCon  & 46.6 & 67.6 & 9.5 & 14.3 & 238   \\
PETS09- &  LP2D  & 40.7 & 70.2 & 9.5 & 16.7 & 319  \\
S2L2 & Proposed & 34.5 & 69.7 & 7.1 & 19.0 & 282 \\\hline
  
  &     MotiCon  & 43.5 & 72.9 & 20.0 & 28.9 & 37 \\
  ETH-   & LP2D  & 40.7 & 73.5 & 15.6 & 26.7 & 41  \\
 Jelmoli & Proposed & 42.3 & 72.8 & 24.4 & 31.1 & 30  \\\hline
  
&    MotiCon  & 18.3 & 77.7 & 1.5 & 74.1 & 72  \\
ETH- &  LP2D & 16.9 & 76.4 & 2.0 & 73.6 & 77  \\
Linthescher & Proposed & 16.7 & 74.2 & 4.6 & 78.7 & 9   \\\hline
  
  &     MotiCon  & 22.8 & 72.9 & 3.8 & 65.4 & 8  \\
  ETH-   & LP2D& 21.4 & 76.3 &3.8 & 65.4 & 10  \\
 Crossing& Proposed & 27.5 & 74.1 & 3.8 & 65.4 & 4  \\\hline
  
&    MotiCon & 11.9 & 70.3 & 0.9 & 69.9 & 74  \\
AVG- &  LP2D & 15.5 & 68.5 & 8.4 & 33.2 & 260  \\
TownCentre & Proposed & 19.3 & 69.0 & 4.4 & 44.7 & 142   \\\hline
  
  &     MotiCon  & 1.0 & 70.3 & 18.8 & 12.5 & 136  \\
  ADL- & LP2D  & 2.9 & 72.2 & 15.6 & 21.9 & 252 \\
 Rundle-1 & Proposed & 26.5 & 71.6 & 28.1 & 28.1 & 33   \\\hline
  
&    MotiCon  & 18.1 & 71.8 & 4.5 & 20.5 & 217  \\
ADL- &  LP2D  & 13.7 & 72.8 & 2.3 & 25.0 & 400 \\
Rundle-3 & Proposed & 39.7 & 72.9 & 11.4 & 34.1 & 33  \\\hline
  
  &     MotiCon  & 38.8 & 70.1 & 0.0 & 11.8 & 36  \\
 KITTI-16 & LP2D  & 35.5 & 72.0 & 0.0 & 11.8 & 47 \\
  & Proposed & 36.9 & 72.6 & 0.0 & 17.6 & 24   \\\hline
  
&    MotiCon & 33.8 & 69.9 & 6.5 & 21.0 & 100 \\
KITTI-19 &  LP2D  & 20.1 & 65.2 & 8.1 & 21.0 & 97 \\
 & Proposed & 26.7 & 66.2 & 6.5 & 29.0 & 70   \\\hline
  
  &    MotiCon  & 18.2 & 72.9 & 0.0 & 29.4 & 74 \\
Venice-1 &  LP2D  & 11.0 & 72.4 & 0.0 & 35.3 & 98 \\
 & Proposed & 22.3 & 73.0 & 0.0 & 41.2 & 4 \\ \hline\hline
  
      \end{tabular}
  \end{center}
    \caption{Detailed result on the 11 sequences of MOTChallenge test, compared to two other methods that use also Linear Programming.}
\label{tab:motcha-seqs}
\end{table}

Finally we show the results on the test set of MOTChallenge in Table \ref{tab:motcha}, where we compare to numerous state-of-the-art trackers. Our method is among the top performing trackers, and contains less false positives than any other method. Note, that we do not use any type of post-processing. Again, it clearly outperforms methods based on Linear Programming (LP2D and MotiCon), thanks to the proposed edge costs.

\begin{table*}[tb]
\begin{center}
 \begin{tabular}{l|  c c c c c c  }
   Method   &  TA & TP & MT & ML  &  IDsw & FP \\ \hline
  NOMT \cite{choiiccv2015}  & {\bf 33.7} & {\bf 71.9} & 12.2 & 44.0 & 442 & 7762 \\
  MHT-DAM \cite{kimiccv2015} & 32.4 & 71.8 & {\bf 16.0} & 43.8 & 435 & 9064 \\
  MDP \cite{xiangiccv2015} & 30.3 & 71.3 & 13.0 & {\bf 38.4} & 680 & 9717 \\  \hline
  SiameseCNN (proposed)  & 29.0 & 71.2 &  8.5 &  48.4 & 639 & \bf{5160} \\ \hline
      LP-SSVM \cite{wangbmvc2015}  & 25.2 & 71.7 & 5.8 & 53.0 & 849 & 8369 \\
    ELP \cite{mclaughlinwacv2015}& 25.0 & 71.2 & 7.5 & 43.8 & 1396 & 7345 \\
    JPDA-m \cite{rezatofighiiccv2015} & 23.8 & 68.2 & 5.0 & 58.1 & {\bf 365} &6373 \\ 
    MotiCon \cite{lealcvpr2014} & 23.1 & 70.9 & 4.7 & 52.0 & 1018 & 10404 \\
        SegTrack \cite{milancvpr2015} & 22.5 & 71.7 & 5.8 & 63.9 & 697 & 7890 \\ 
    LP2D (baseline)  & 19.8 & 71.2 & 6.7 & 41.2 & 1649 & 11580  \\
    DCO-X \cite{milanpami2016} & 19.6 & 71.4 & 5.1 & 54.9 & 521 & 10652 \\
    CEM \cite{milanpami2014} & 19.3 & 70.7 & 8.5 & 46.5 & 813 & 14180 \\
	RMOT \cite{yoonwacv2015} & 18.6 & 69.6 & 5.3 & 53.3 & 684 & 12473 \\
	SMOT \cite{dicleiccv2013} & 18.2 & 71.2 & 2.8 & 54.8 & 1148 & 8780 \\
	ALExTRAC \cite{bewleyicra2016} & 17.0 & 71.2 & 3.9 & 52.4 & 1859 & 9233 \\
	TBD \cite{geigerpami2014} & 15.9 & 70.9 & 6.4 & 47.9 & 1939 & 14943 \\
	TC-ODAL \cite{baecvpr2014} & 15.1 & 70.5 & 3.2 & 55.8 & 637 & 12970 \\
	DP-NMS \cite{pirsiavashcvpr2011} & 14.5 & 70.8 & 6.0 & 40.8 & 4537 & 13171 \\
	LDCT \cite{soleraiccv2015} & 4.7 & 71.7 & 11.4 & 32.5 & 12348 & 14066 \\
    \end{tabular}
  \end{center}
  \vspace{-0.5cm}
    \caption{Results on the MOTChallenge test set.}
\label{tab:motcha}
\end{table*}

%We also include in the measures the provided Average Rank, which englobes all the measures present in the benchmark. As we can see, we are ranked 3rd best tracker among all published work. 

\section{Conclusions}

\par In this paper we have presented a two-stage learning based approach to associate detections within the context of pedestrian tracking. In a first pass, we create a multi-dimensional input blob stacking  image and optical flow information from to the two patches to be compared; these data representation allows the following Siamese convolutional neural network to learn the relevant spatio-temporal features that allow distinguishing whether these two pedestrian detections belong to the same tracked entity. These local features are merged with some contextual features by means of a gradient boosting classifier yielding to a unified prediction.

\par In order to highlight the efficiency of the proposed detection association technique, we use a modified linear programming based tracker \cite{zhangcvpr2008} to link the proposed correspondences and form trajectories. The complete system is evaluated over the standard MOTChallenge dataset \cite{motchallenge:arxiv:2015}, featuring enough data to ensure a satisfactory training of the CNN and a thorough and fair evaluation. When comparing the proposed results with the state-of-the-art, we observe that a simple linear programming tracker fed with accurate information reaches comparable performance than other more complex approaches.

\par Future research within this field involve applying the proposed approach to more generic target tracking, leveraging already trained models and extending the second stage classifier to deal with more complex contextual features, e.g. social forces \cite{lealiccv2011}. Evaluation of the proposed architecture over on datasets is currently under investigation.

\bibliographystyle{ieee}
\footnotesize
\bibliography{cvpr2016,refs-alt}

\end{document}